\documentclass[letterpaper, 10 pt, conference]{ieeeconf}

\IEEEoverridecommandlockouts

\usepackage{cite}
\usepackage{amsmath,amssymb,amsfonts}
\usepackage[utf8]{inputenc}
\usepackage[T1]{fontenc}    
\usepackage{algorithmic}
\usepackage{graphicx}
\usepackage{textcomp}
\usepackage{amsmath,bm}
\usepackage{subcaption}
\usepackage{booktabs}
\let\labelindent\relax
\usepackage{enumitem}
\usepackage{multirow}
\usepackage{xcolor}
\usepackage{colortbl}
\usepackage[colorlinks=true, urlcolor=red, breaklinks=true]{hyperref}
\usepackage{breakurl}
\usepackage{url}

\usepackage[skip=0pt]{caption} 
\def\BibTeX{{\rm B\kern-.05em{\sc i\kern-.025em b}\kern-.08em
    T\kern-.1667em\lower.7ex\hbox{E}\kern-.125emX}}

\renewcommand{\baselinestretch}{.98}

\newtheorem{remark}{Remark}

\begin{document}

\title{Robust Online Calibration for UWB-Aided Visual-Inertial Navigation with Bias Correction}

\author{%
	Yizhi Zhou, Jie Xu, Jiawei Xia, Zechen Hu, Weizi Li, Xuan Wang 
 \thanks{Y. Zhou, J. Xia, Z. Hu and X. Wang are with the
		Department of Electrical and Computer Engineering, George Mason
		University. J. Xu is with the Department of Electrical and Computer Engineering, University of California, Riverside. W. Li is with the Department of Electrical Engineering and Computer Science, University of Tennessee, Knoxville.            
  }
  \thanks{This research is supported by NSF IIS-$2153426$ and ECCS-$2332210$.}
}

\maketitle

\begin{abstract}
This paper presents a novel robust online calibration framework for Ultra-Wideband (UWB) anchors in UWB-aided Visual-Inertial Navigation Systems (VINS). Accurate anchor positioning, a process known as calibration, is crucial for integrating UWB ranging measurements into state estimation. While several prior works have demonstrated satisfactory results by using robot-aided systems to autonomously calibrate UWB systems, there are still some limitations: 1) these approaches assume accurate robot localization during the initialization step, ignoring localization errors that can compromise calibration robustness, and 2) the calibration results are highly sensitive to the initial guess of the UWB anchors' positions, reducing the practical applicability of these methods in real-world scenarios. Our approach addresses these challenges by explicitly incorporating the impact of robot localization uncertainties into the calibration process, ensuring robust initialization. To further enhance the robustness of the calibration results against initialization errors, we propose a tightly-coupled Schmidt Kalman Filter (SKF)-based online refinement method, making the system suitable for practical applications. Simulations and real-world experiments validate the improved accuracy and robustness of our approach.
\end{abstract}

\section{Introduction}
Visual-inertial navigation system (VINS) is favored in robot state estimation due to its accuracy, reliability, and lightweight design \cite{OP2020, VM}. Nevertheless, VINS suffers from cumulative drift due to inherent limitations in visual-based localization methods. While GPS provides a natural solution for external localization information in outdoor environments, its reliance on open spaces makes it unsuitable for use in GPS-denied or indoor settings. To address this, many recent works incorporate Ultra Wideband (UWB) measurements into the VINS to leverage the global observation provided by UWB for better localization performance \cite{Jia2022, Hc2024, NTM2021}. 

Specifically, UWB-aided VINS utilizes ranging measurements between the robot and the UWB anchor to enhance robot state estimation. Accurate robot localization requires precise knowledge of these UWB anchor positions, a process known as UWB calibration. Many previous studies have proposed self-calibration, where a robot autonomously calibrates the anchor positions using the geometric relations between its position and the anchor positions, subsequently integrating these calibration results into the VINS \cite{Hj2023, BJ2021}. Most UWB self-calibration methods typically involve two key steps: an initialization phase to estimate coarse anchor positions, followed by a refinement phase to improve these estimates. These methods
fall into two categories: (i) loosely-coupled methods, where calibration is handled by a separate estimator independent of robot localization, and (ii) tightly-coupled methods, where both the robot state and UWB state are jointly estimated within a single framework. Tightly-coupled methods offer greater flexibility \cite{NT2020} and outperform loosely-coupled approaches in both anchor calibration and robot localization \cite{Hc2024}.

However, during the initialization phase, many existing works implicitly assume that the localization of robots is accurate. This assumption neglects the impact of localization uncertainties, which can lead to poor initial estimates of UWB anchor positions.
Additionally, the performance of tightly-coupled calibration systems are heavily sensitive to the accuracy of initial UWB parameters \cite{NT2022}. In practice, obtaining precise initial guesses is challenging, potentially degrading system performance and, in some cases, causing system failure.

\begin{figure}[t]
	\centering
	\includegraphics[width=0.48\textwidth]{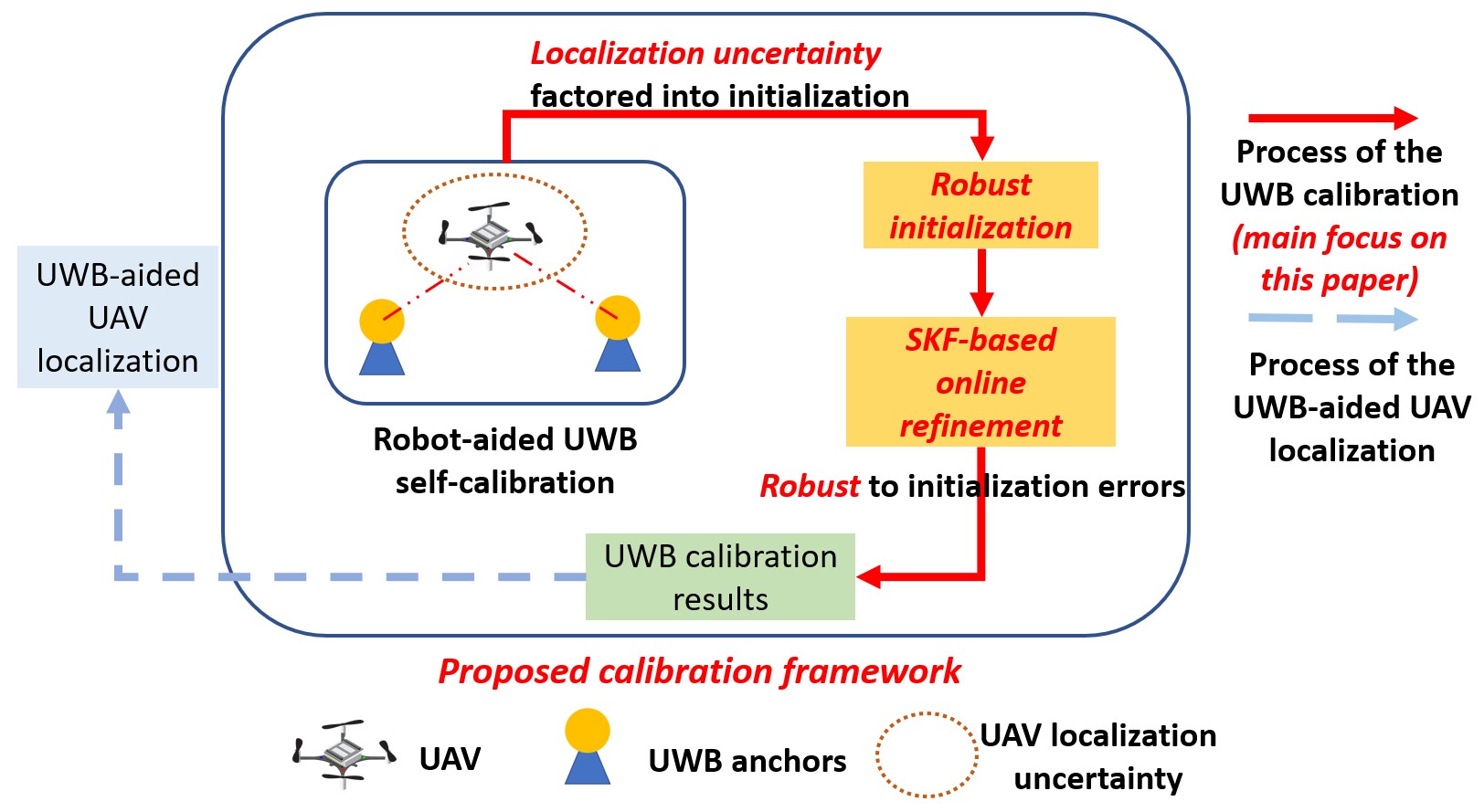}
        \vspace{-.5ex}	
        \caption{System Overview: The proposed robot-aided UWB calibration system contains two stages: A robust initialization stage and a SKF-based refinement stage.}
	\label{Fig_exp_diag}
	\vspace{-2ex}
\end{figure}

\subsubsection*{Statement of Contribution} To address the challenges identified in UWB calibration, we propose a robust calibration method for improving both the performance and reliability of existing calibration frameworks. The main contributions of this paper are as follows:
\begin{itemize}[leftmargin=*]
    \item We propose a novel UWB initialization method that explicitly accounts for uncertainty in robot localization during the initialization process, improving both the performance and robustness of calibration in the presence of various
    robot localization uncertainties.
    \item We show, through experiments and analysis, that the initialization phase of UWB calibration can significantly impact the overall performance of the calibration system. We propose a tightly-coupled Schmidt Kalman Filter (SKF)-based online calibration method to reduce the influence of initialization errors on the calibration process.
    \item We perform extensive simulations and experiments to validate the performance of the initialization and calibration methods compared with baselines.
\end{itemize}

\section{Related Works}
UWB measurements have been widely utilized to enhance robot localization in numerous studies. In early researches \cite{WC2017, PF2017, ZJ2022, YB2021, SSL2021}, anchor positions were calibrated offline and used as prior knowledge for drift-free state estimation. While these methods effectively reduce the drift in VINS, performing offline calibration becomes challenging in dynamic or large-scale environments.

Several studies have explored the use of raw UWB measurements and robot localization information for self-calibration \cite{ZS20222}. Typically, self-calibration of UWB involves two key steps: an initialization phase to estimate coarse anchor positions, followed by a refinement phase to improve the initial estimates. 
In \cite{Jia2022}, a consistent visual-inertial-ranging-odometry system is proposed, where the anchor positions are initially estimated by solving a least squares problem and are then jointly optimized with the VINS state in a tightly-coupled manner. Refs. \cite{NT2022, BJ2021} observed that the robot's trajectory can significantly influence the calibration process, and design path-planning algorithms to compute an optimal trajectory for improved calibration accuracy. The studies in \cite{BJ2021, ZW2021} notice that the ranging measurement can be deteriorated by additional distance-dependent biases due to the signal block in cluttered environments. Even though the anchor position is precisely calibrated, the inconstant bias may introduce additional errors. Therefore, the bias term of the ranging measurements are introduced to improve the calibration. In \cite{JS2023}, a fully distributed calibration framework is introduced to initialize large-scale UWB networks, significantly reducing computational complexity on the robot's side. 
For the purpose of robust calibration, multiple studies have pointed out a crucial fact that the calibration results are sensitive to the initial guess and hence aim to enhance the initial guess for robust calibration performance \cite{DGS2023, NT2022, Hc2024}. To avoid the non-line-of-sight (NLOS) problem, an object detection-based calibration method is proposed in \cite{Hc2024} that directly uses visual measurements to initialize the anchor positions.

Nevertheless, none of the above mentioned works consider the fact that the robustness of the calibration is highly sensitive to the accuracy of the robot's localization and the initialization values. Therefore, these approaches still have the following limitations: 1) Anchor initialization is based on the assumption of accurate robot positioning, which compromises the robustness of the estimated anchor positions; and 2) Jointly estimating the robot state and anchor positions, if the anchor has an inaccurate initial guess, can significantly degrade the robot's localization, leading to further inaccuracies in the calibration process.

\section{Preliminaries}
\subsection{Problem Formulation}
Consider a robot in a 3-D environment equipped with a visual-inertial sensor for ego-motion measurement and a UWB tag for distance measurement. We first define the IMU and camera frames as $I$ and $C$, respectively, while $G$ represents the global frame. There are $M$ position-unknown UWB, denoted as ${^G\mathbf p_{a_i}}, i=\{1,2,...,M\}$, to be estimated in our setup.

The UWB tag on the robot can provide the ranging measurement 
between the robot and the UWB anchor $i$ at timestep $t_k$, denoted as $d_{i,k}$. Given the UAV's pose $({^{I_k}_G{\mathbf R}}, ^G\mathbf p_{I_k})$ and the anchor position ${^G\mathbf p_{a_i}}$, the ranging measurement is described by the following model:
\begin{align}\label{eq_uwb}
d_{i,k}&=\beta_i(\|^G\mathbf p_{I_k}+{^{I_k}_G{\mathbf R}}^\top{^{I}\mathbf p_T}-{^G\mathbf p_{a_i}}\|+\mathbf n_u)+\gamma_i
\end{align}
where $\mathbf n_u\sim \mathcal N(0, \mathbf Q_d)$, and ${^{I}\mathbf p_T}$ represents UAV tag's position in its IMU frame, which can be easily calibrated offline. Two bias terms $\beta_i$ and $\gamma_i$, firstly introduced in \cite{BJ2021}, are included to better reflect, in practice, the difference of UWB sensor reading and the true distance.
It is important to note that $\beta_i$ and $\gamma_i$ are path-dependent biases \cite{BJ2021}, designed to capture the position dependency of the range error.

In the UWB-aided VINS, the UWB should be calibrated at first, so that the UWB ranging measurements can participate the estimation of robot position. This paper primarily focuses on UWB calibration rather than robot localization. The objective is to use a UAV equipped with a visual-inertial sensor to automatically calibrate the UWB, i.e., estimate $({^G\mathbf p_{a_i}}, \beta_i,\gamma_i)$, while ensuring robust and consistent calibration results by accounting for the uncertainty in robot localization.

\subsection{System Overview}

We propose a two-phase calibration scheme for the UWB-aided VINS as shown in Fig. ~\ref{Fig_exp_diag}. It is important to note that the method of this paper focuses solely on the robot-aided calibration phase, rather than UWB-aided robot localization.
In particular, the proposed calibration system consists of two phases: a robust initialization phase ({\color{red} Sec. \ref{RI}}) and a SKF-based online refinement phase ({\color{red} Sec. \ref{SKF}}). In the first phase, 
the objective is to find an initial guess of the anchor position ${^G\mathbf p_{a_i}}$ and the unknown bias parameters $(\beta_i,\gamma_i)$. The VINS runs normally after the visual-inertial initialization is completed. Once the UWB tag receives a ranging measurement and the ID from any of the anchors, the ranging measurement and the robot pose estimate will be stored for initialization. Since the robot localization results are directly used to initialize the UWB, the uncertainty of robot localization also affects the UWB estimation. To ensure robust estimation, we incorporate VINS uncertainty into the initialization process. Once the initialization phase is completed, the system transfers to the online refinement phase to improve the initial UWB guess in the second stage. 


\section{VINS-Aided Calibration System} \label{System}
In this section, we extend the standard Multi-State Constraint Kalman Filter (MSCKF)-based VINS \cite{MSCKF} framework to incorporate additional calibration for the UWB system.

\subsection{System State and Model}
The state vector at each timestep $t_k$, denoted as $\mathbf x_{k}$,
is defined by
\begin{align}
\mathbf x_k&\triangleq \begin{bmatrix}
\mathbf x_{r_k}^\top& \mathbf x_{U_k}^\top
\end{bmatrix}
=\begin{bmatrix}
\mathbf x_{I_k}^\top& \mathbf x_{C_k}^\top & \mathbf x_{U_k}^\top
\end{bmatrix}^\top\nonumber\\
\mathbf x_{I_k}&=\begin{bmatrix}
{^{I_k}_G{\mathbf q}^\top}& \mathbf b_{g}^\top & {^G \mathbf v_{I_k}^\top}& \mathbf b_{a}^\top& {^G \mathbf p_{I_k}^\top}
\end{bmatrix}^\top\nonumber\\
\mathbf x_{C_k}&=\begin{bmatrix}{^{C_k}_G{\mathbf q}^\top}&{^G \mathbf p_{C_k}^\top}&...&{^{C_{k-N}}_G{\mathbf q}^\top}&{^G \mathbf p_{C_{k-N}}^\top}
\end{bmatrix}^\top\nonumber\\
\mathbf x_{U_k}&=\begin{bmatrix}
{^G \mathbf p_{a}^\top}& \beta & \gamma
\end{bmatrix}^\top
\end{align}
where $\mathbf x_{r_k}$ denotes the active MSCKF state.
The $\mathbf x_{I_k}$ is the IMU state including IMU's orientation, position, velocity, and biases; $\mathbf x_{C_k}$ is the clone of historical IMU poses when features are observed by camera; $\mathbf x_{C_k}$ denotes the clone of the historical cameras states. $\mathbf x_{U_k}$ denotes the UWB state which contains the anchor position ${^G \mathbf p_{a_k}^\top}$ the UWB model parameters $(\beta, \gamma)$. For simplicity, we consider only a single UWB anchor in our analysis; Nevertheless, this framework can be easily extended to multiple anchors.
To represent the system's covariance matrix more clearly, we introduce the following partitioning for the covariance
\begin{align}\label{eq_state_cov}
\mathbf P_{k|k}&= \begin{bmatrix}
\mathbf P_{II_{k|k}}&\mathbf P_{IC_{k|k}}&\mathbf P_{IU_{k|k}}\\
\mathbf P_{IC_{k|k}}^\top&\mathbf P_{CC_{k|k}}&\mathbf P_{CU_{k|k}}\\
\mathbf P_{IU_{k|k}}^\top&\mathbf P_{CU_{k|k}}^\top&\mathbf P_{UU_{k|k}}
\end{bmatrix}
\end{align}
where $\mathbf P_{II_{k|k}}\in \mathbb R^{15\times 15}$ is the covariance of the IMU state, $\mathbf P_{CC_{k|k}}\in \mathbb R^{6N\times 6N}$ is the covariance of the camera estimate, and $\mathbf P_{UU_{k|k}}\in \mathbb R^{5\times 5}$ is the covariance of the UWB state; $\mathbf P_{IC_{k|k}}, \mathbf P_{IU_{k|k}}, \mathbf P_{CU_{k|k}}$ denotes the corresponding cross-correlations. The covariance matrix will be initially constructed upon completion of the initialization step, as described in ({\color{red}Sec. \ref{RI}}).

The state $\mathbf x_k$ will be propagated forward with IMU's linear velocity and acceleration measurements based on the IMU kinematic model\cite{SK2018}. To propagate the state covariance matrix, we linearize the IMU kinematics and compute the state transition matrix $\bm\Phi(t_{k+1},t_k)$ \cite{SK2018}. With notation \eqref{eq_state_cov}, the corresponding state covariance can be propagated as
\begin{align}\label{eq_cov_pro}
\mathbf P_{k+1|k}&= \begin{bmatrix}
\mathbf P_{II_{k+1|k}}&\mathbf P_{IC_{k+1|k}}&\mathbf P_{IU_{k+1|k}}\\
\mathbf P_{IC_{k+1|k}}^\top&\mathbf P_{CC_{k|k}}&\mathbf P_{CU_{k|k}}\\
\mathbf P_{IU_{k+1|k}}^\top&\mathbf P_{CU_{k|k}}^\top&\mathbf P_{UU_{k|k}}
\end{bmatrix}
\end{align}    
where $\mathbf P_{II_{k+1|k}}$ is the propagated IMU covariance.
$\mathbf P_{IC_{k+1|k}} = \bm\Phi(t_{k+1}, t_k)\mathbf P_{IC_{k|k}}$, and $\mathbf P_{IU_{k+1|k}} = \bm\Phi(t_{k+1}, t_k)\mathbf P_{IU_{k|k}}$.

\subsection{Measurement Update Model}
\noindent{\textbf{Camera update model:}}
When a static landmark of the environment, denoted as ${^{G}\mathbf p_f}$, is tracked by the camera at timestep $t_k$, the corresponding feature measurement can be obtained through the following model
\begin{align}
\mathbf z_{C,k}&=\Lambda({^{C_k}\mathbf p_f})+\mathbf n_c\nonumber\\
{^{C_k}\mathbf p_f}&={^{C}_{I} \mathbf R}{^{I_k}_G \mathbf R}({^{G}\mathbf p_f}-{^{G}\mathbf p_{I_k}})+{^{C}\mathbf p_{I}}
\end{align}
where $\mathbf n_c\sim \mathcal N(0, \mathbf Q_c)$ is the white Gaussian noise with covariance $\mathbf Q_c$. The ${^{C_k}\mathbf p_f}$ is the landmark position in the camera frame, and the projection function $\Lambda(\cdot)$ is defined as $\Lambda(\begin{bmatrix}x&y&z\end{bmatrix}^\top)=\begin{bmatrix}x/z&y/z
\end{bmatrix}^\top$. 
By linearizing the measurement equation, the camera measurements can be incorporated into the EKF update, as detailed in \cite{MSCKF}.

\noindent{\textbf{UWB update model:}}
Once the robot receives a ranging measurement from the anchor, this measurement will be used for state update. Given the measurement model \eqref{eq_uwb}, we linearize it at the current estimate $\hat{\mathbf x}_{I_k}$ as
\begin{align}
\Tilde{d}_{k}&= \mathbf H_{I_k} \Tilde{\mathbf x}_{I_k}+\mathbf H_{U_k}\Tilde{\mathbf x}_{U_k}+\mathbf n_u
\end{align}
Since our formulation and derivation assume only a single anchor, we simplify the notation by omitting the subscript $i$ in the variable $d_{i,k}$ in the following derivation. The Jacobian corresponding to the IMU state, denoted as $ \mathbf{H}_{I_k} $, and the Jacobian corresponding to the UWB state, which includes  $(^G\mathbf{p}_{a}, \beta, \gamma) $, denoted as $\mathbf{H}_{U_k}$, can be computed as:
\begin{align}\label{eq_uwb_j}
\mathbf{H}_{I_k}&=\mathbf H_p\begin{bmatrix}
-\lfloor{^I_G\hat{\mathbf R}^\top} {^I\mathbf p_T}\times\rfloor & \mathbf 0_{3\times 9} & \mathbf I_3
\end{bmatrix}\nonumber\\
\mathbf{H}_{U_k}&=\begin{bmatrix}
-\mathbf H_p & \mathbf H_{\beta} & 1
\end{bmatrix}
\end{align}
where the matrices $\mathbf H_p$ and $\mathbf H_{\beta}$ can be computed as
\begin{align}
\mathbf H_p&=\frac{\hat\beta\left(^G\hat{\mathbf p}_{I_k}+{^{I_k}_G\hat{\mathbf R}^\top}{^{I}\mathbf p_T}-{^G\hat{\mathbf p}_a}\right)^\top}{\|^G\hat{\mathbf p}_{I_k}+{^{I_k}_G\hat{\mathbf R}^\top}{^{I}\mathbf p_T}-{^G\hat{\mathbf p}_a}\|}\nonumber\\
\mathbf H_{\beta}&=\|^G\hat{\mathbf p}_{I_k}+{^{I_k}_G\hat{\mathbf R}^\top}{^{I}\mathbf p_T}-{^G\hat{\mathbf p}_a}\|
\end{align}


\section{Robust Initialization by Stochastic Optimization}\label{RI}

To integrate the ranging measurement into the navigation system, an initial estimate of the UWB anchor position ${^G\mathbf p_{a}}$ and the model parameters $(\beta,\gamma)$ is required for the anchor. To simplify the initialization during implementation, we temporarily fix the value of $\beta = 1$ at this stage, and update it in the next refinement stage. We assume that the robot's state, as obtained through the VINS, along with ranging measurements, are readily accessible during the initialization phase, stored within a time window of length $m$. 
\begin{remark}
The parameter $\beta$ in the UWB model \eqref{eq_uwb} is typically very close to $1$ in actual test results. In fact, many papers simply assume this value to be $1$ for practical use \cite{ZW2021, Jia2022}. Therefore, during the initialization stage, we set $\beta$ to 1 as an initial guess. It is then refined in the subsequent stage (see refinement stage in {\color{red}{Sec. IV}}).
\end{remark}

\subsection{Robust Initialization Formulation}
Building on the previous analysis, the uncertainty in the UAV's pose estimate can significantly affect UWB calibration and must be considered for robust performance. Although VINS drift is small during the initialization phase, neglecting this uncertainty can lead to inconsistencies in the estimator and degrade initialization accuracy. To address this, we propose a robust initialization method that explicitly integrates UAV pose uncertainty into the calibration process. Specifically, we formulate the following stochastic robust approximation problem \cite[Sec.~6.4]{boyd2004}
 to estimate the UWB state $({^G\mathbf p_{a}}, \gamma)$, treating the robot's state as a stochastic variable
\begin{align}\label{eq_op}
\min_{^G\mathbf p_a,\gamma}\sum_{k=1}^{m}\mathbb{E}_{\mathbf x_{I_k}}(\| d_{k}- h({\mathbf x}_{I_k}, {\mathbf x}_{U})\|^2)
\end{align}
\textit{where the above expectation $\mathbb{E}(\cdot)$ is taken over the instantiations of all possible robot (IMU) state ${\mathbf x}_{I_k}$}. This accounts for the fact that ${\mathbf x}_{I_k}$ is a random variable subject to uncertainties. $ d_{k}$ is the collected ranging measurement. $h(\cdot)$ is the UWB measurement function, referenced in \eqref{eq_uwb}. ${\mathbf x}_{U_k}=({^G {\mathbf p}_{a_i}}, {\gamma}_i)$ is the UWB state to be estimated. Since the cost in \eqref{eq_op} involves an expectation which cannot be directly solved, a common approach is to approximate the expectation and reformulating the problem into a regular optimization formulation. In particular, we linearize the measurement function $\mathbf h(\cdot)$ at the current UAV's state estimate $\hat{\mathbf x}_{I_k}$ using the first-order Taylor expansion to approximate the cost function \eqref{eq_op} as
\begin{align}
&\sum_{k=1}^m\mathbb{E}_{\mathbf x_{I_k}}(\| d_{k}-h({\mathbf x}_{I_k}, {\mathbf x}_{U})\|^2)\approx\nonumber\\
&\sum_{k=1}^m \left( \| d_{k}-h(\hat{\mathbf x}_{I_k}, {\mathbf x}_{U})\|^2+
\text{trace}(\mathbf H_{I_{k}} \mathbf P_{II_k}\mathbf H_{I_{k}}^\top)\right)
\end{align}
where the jacobian $\mathbf H_{I_{k}}$ is computed in \eqref{eq_uwb_j}, and the term $\text{trace}(\mathbf H_{I_{k}}^\top \mathbf P_{II_k}\mathbf H_{I_{k}})$ denotes the contribution of the UAV's pose uncertainty to the UWB estimation. By incorporating this term into the cost function, the uncertainty of the UAV's pose is explicitly considered in the optimization process, leading to a more robust initialization. This approximated cost function can then be efficiently minimized using iterative methods such as Gradient Descent or Gauss-Newton. A more detailed derivation of this approximation process is provided in {\color{red}{App. A}} of the supplementary material \cite{zzSPLY2025}.

\begin{figure}[t]
	\centering
	\includegraphics[width=0.48\textwidth]{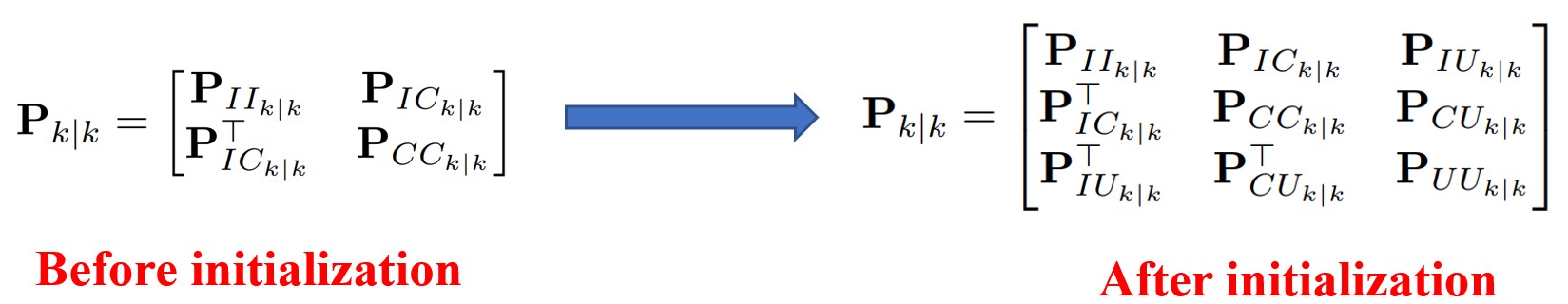}
        \vspace{-.5ex}	
        \caption{To ensure a consistent state estimation, we need to compute the covariance of the UWB state and its cross-correlations with the existing MSCKF state. This step is crucial for maintaining the integrity of the filter, as it properly accounts for the uncertainties associated with the newly introduced UWB state.}
	\label{Fig_cinit}
	\vspace{-0.5ex}
\end{figure}

This coarse estimate will serve as an initial guess for the tightly-coupled online refinement approach described in the following section. Before the refinement step, the state covariance has to be initialized in accordance with the structure of \eqref{eq_state_cov}. In particular, we have to initialize the covariance of the UWB state and its correlations with the existing state variables as shown in Figure \ref{Fig_cinit}. We adopt a method very similar to the "state variable initialization" method described in \cite{OP2020}. We first define a state $\mathbf X_{I}$ that contains the IMU state from timestep $k=0$ to $k=m$ as
\begin{align}
\mathbf X_{I}=\begin{bmatrix}
\mathbf x_{I_0}^\top & \mathbf x_{I_1}^\top & \cdots &\mathbf x_{I_m}^\top
\end{bmatrix}^\top
\end{align}
and then stack all the available UWB measurements $\mathbf d=\begin{bmatrix}
\mathbf d_0 & \cdots &\mathbf d_m
\end{bmatrix}^\top$ to construct a stacked measurement model as
\begin{align}\label{eq_meas_stack}
\mathbf d&=\mathbf h(\mathbf X_{I}, \mathbf x_{U})\nonumber\\
\mathbf h(\mathbf X_{I}, \mathbf x_{U})&=[h(\mathbf x_{I_0}, \mathbf x_{U}), ..., h(\mathbf x_{I_m}, \mathbf x_{U})]^\top
\end{align}
where $h(\cdot)$ satisfies the UWB measurement model \eqref{eq_uwb}.
Then we linearize the stacked measurement model, which yields
\begin{align}\label{eq_ln_init}
\widetilde{\mathbf d}&=\begin{bmatrix}
\mathbf H_A & \mathbf H_B 
\end{bmatrix}\begin{bmatrix}
\widetilde{\mathbf X}_{I} \\ \widetilde{\mathbf x}_{U}
\end{bmatrix}+\bar{\mathbf n}_u
\end{align}
where $\bar{\mathbf n}_u$ represents the stacked noise vector, and $\mathbf H_A$ and $\mathbf H_B$ are the corresponding measurement jacobians given by
\begin{align}
\mathbf H_A&=\text{diag}(\mathbf H_{I_0}, \cdots, \mathbf H_{I_k})\nonumber\\
\mathbf H_B&=\text{diag}(\mathbf H_{U_0}, \cdots, \mathbf H_{U_k})
\end{align}
We then decompose the linearized system \eqref{eq_ln_init} into two subsystems using QR decomposition, expressed as:
\begin{align}
\begin{bmatrix}
\widetilde{\mathbf d}_{1} \\ \widetilde{\mathbf d}_{2}
\end{bmatrix}&=\begin{bmatrix}
\mathbf H_{A_1} & \mathbf H_{A_2} \\ \mathbf H_{B_1} & \mathbf 0
\end{bmatrix} \begin{bmatrix}
\widetilde{\mathbf X}_{I} \\ \widetilde{\mathbf x}_{U}
\end{bmatrix} + \begin{bmatrix}
\bar{\mathbf n}_{u_1}\\ \bar{\mathbf n}_{u_2}
\end{bmatrix}
\end{align}
Then, the covariance of the UWB state and its correlations to the existing state $\mathbf x_{I_k}$ and $\mathbf x_{C_k}$ can be computed and used to augmented to the current covariance following \cite{OP2020}.

\subsection{FIM-based Analysis}\label{FIM}
This section presents an analysis of the effectiveness and performance of the proposed robust initialization using the Fisher Information Matrix (FIM) \cite{WZ2008}, which quantifies the information contained in the observed data about the unknown parameter being estimated. Mathematically, it is defined as the expectation of the outer product of the gradient of the log-likelihood function \cite{WZ2008}:
\begin{align}\label{eq_Fd}
\mathbf{F} = \mathbb{E} \left[ \left( \frac{\partial \ell( \mathbf d;\mathbf{x}_U)}{\partial \mathbf{x}_U} \right) \left( \frac{\partial \ell(\mathbf d;\mathbf{x}_U)}{\partial \mathbf{x}_U} \right)^\top \right], 
\end{align}
where the definition of $\mathbf d$ is provided in \eqref{eq_meas_stack} and $\mathbf x_U$ denotes the UWB state to be estimated; $\ell(\mathbf d;\mathbf{x}_U)$ represents the log-likelihood function of the measurement model. To calculate this log-likelihood function $\ell(\mathbf d;\mathbf{x}_U)$, we first formulate the corresponding likelihood function, denoted as $\mathbf p(\mathbf d; \mathbf{x}_U)$, based on the stacked measurement model \eqref{eq_meas_stack}, as follows
\begin{align}\label{eq_likh}
\mathbf p(\mathbf d; \mathbf{x}_U)&=\frac{1}{(2\pi)^{\frac{N}{2}} \det(\Sigma)^{\frac{1}{2}}} \exp\left( {\left( {\mathbf d} - \mathbf h \right)}^\top \bm\Sigma^{-1} \left( {\mathbf d} - \mathbf h \right)\right)
\end{align}
where $\mathbf\Sigma=\text{diag}(\Sigma_0,\cdots, \Sigma_k, \cdots\Sigma_m)$ denotes the covariance which quantifies the overall system uncertainty. In most existing works \cite{Hj2023, NT2022}, the term $\Sigma_k$ is typically regarded as the measurement covariance $\mathbf Q_d$. However, as previously analyzed, uncertainty arises not only from measurement noise but also from the robot’s localization error. Therefore, it is crucial to explicitly account for localization uncertainty in the covariance term $\mathbf\Sigma$ during initialization. In our approach, we model the localization uncertainty $\mathbf{P}_{II_k}$  as additional measurement noise, leading to the modified covariance formulation:
\begin{align}
\Sigma_k=\mathbf Q_d+ \mathbf H_{I_k}\mathbf P_{II_k}\mathbf H_{I_k}^\top
\end{align}
It is important to note that the covariance $\Sigma_k$ is also a function of the UWB estimate $\mathbf x_U$, as the value of the jacobian $\mathbf {H}_{I_k}$ depends on the UWB state $\mathbf x_U$, as derived in {\color{red} Sec. IV-B} of the paper. Therefore, both the mean and covariance of $\mathbf d$ are functions of $\mathbf x_U$. Such a distribution can be written as $\mathbf d\sim\mathcal N(\mathbf h(\mathbf x_U), \bm\Sigma(\mathbf x_U))$, which is referred to as the \textit{General Gaussian Distribution} \cite[Section~3.9]{Kay1993}. For this \textit{General Gaussian Distribution}, there is also a general expression of the FIM according to \cite[Section~3.9]{Kay1993}, where the $(i,j)$-th element of $\mathbf F$ is given as
\begin{align}\label{eq_FIM}
    \mathbf F_{ij}&=\left(\frac{\partial \mathbf h(\mathbf x_U)}{\partial \mathbf x_U^i}\right)^\top \bm \Sigma(\mathbf x_U^i)^{-1}\left(\frac{\partial \mathbf h(\mathbf x_U)}{\partial \mathbf x_U^j}\right)+\nonumber\\
    &\frac{1}{2}\textbf{trace}\left(\bm \Sigma(\mathbf x_U^i)^{-1}\frac{\partial \mathbf h(\mathbf x_U)}{\partial \mathbf x_U^i}\bm \Sigma(\mathbf x_U^i)^{-1}\frac{\partial \mathbf h(\mathbf x_U)}{\partial \mathbf x_U^j}\right)
\end{align}
where $\mathbf x_U^i$ and $\mathbf x_U^j$ denotes the $i$-th and $j$-th element of the state vector $\mathbf x_U$, respectively. A more detailed derivation and analysis of the FIM is provided in {\color{red} App. B} of the supplementary material \cite{zzSPLY2025}.


\begin{remark}
Due to that $\bm\Sigma$ is a function of $\mathbf x_U$, the second term of (\ref{eq_FIM}) is non-zero, unlike the FIM with a zero-mean Gaussian assumption. Since $\text{det}(\mathbf F)$ is inversely proportional to the uncertainty level, from \eqref{eq_FIM}, we can make a simple inference about the algorithm's performance: 1) As the uncertainty in robot localization $\mathbf x_{I_k}$ increases, $\text{det}(\mathbf F)$ decreases, leading to reduced initialization accuracy; 2) Static robot motion or motion constrained to a single plane, such as movement limited to the xy-plane, can cause $\text{det}(\mathbf F)$ to become singular, making it unsolvable.
\end{remark}

\section{SKF-based Online Refinement}
After obtaining an initial estimate of the unknown UWB parameters, the next step is to refine this estimate, in the following represented as $\mathbf{x}_U=\{{^G\mathbf p_{a_i}}, \beta_i,\gamma_i\}$, to obtain a more accurate calibration.
In this section, we propose a SKF-based estimator built upon the previously discussed MSCKF-based calibration system ({\color{red}{Sec. \ref{System}}}), to perform robust refinement in a tightly-coupled fashion.


\subsection{Issues of Standard EKF-based Update}\label{NA}
Before introducing the proposed SKF-based framework, we first analyze the issues of the standard EKF-based approach and provide numerical tests to demonstrate it, which motivates the proposed SKF-based approach.
According to the standard EKF,\footnote{Throughout this paper, $\hat {\mathbf x}_{k+1|k}$, $\hat {\mathbf x}_{k+1|k+1}$ denotes the prior and posterior estimate at time $t_{k+1}$, respectively. The operator $\boxplus$ denotes the EKF update process, where the state estimate is corrected using the update $\delta \mathbf{x}$ as $\hat{\mathbf x}_{k+1|k+1} = \hat{\mathbf x}_{k+1|k} \boxplus \delta \mathbf{x}$} the system state is updated by either camera measurements or UWB ranging measurements as follow:
\begin{align}
\hat{\mathbf x}_{k+1|k+1}&=\hat{\mathbf x}_{k+1|k}\boxplus\mathbf K \mathbf r\nonumber\\
{\mathbf P}_{k+1|k+1}&={\mathbf P}_{k+1|k}-\mathbf K \mathbf S \mathbf K^\top
\end{align}
where $\mathbf r$ denotes the measurement residual; The Kalman gain $\mathbf K$ and the innovation $\mathbf S$ can be computed as
\begin{align}
\mathbf K&=\begin{bmatrix}\mathbf K_r\\\mathbf K_U\end{bmatrix}= {\mathbf P}_{k+1|k}-\begin{bmatrix}\mathbf H_r^\top\\\mathbf H_U^\top\end{bmatrix}\mathbf S^{-1}\\
\mathbf S&=\begin{bmatrix}\mathbf H_r&\mathbf H_U\end{bmatrix} {\mathbf P}_{k+1|k}
\begin{bmatrix}\mathbf H_r^\top\\ \mathbf H_U^\top\end{bmatrix}+\mathbf Q
\end{align}
where we partition the gain $\mathbf K$ into two parts $\mathbf K_r$ and $\mathbf K_U$, denoting the Kalman gain for the active MSCKF state $\mathbf x_r$ and the UWB state $\mathbf x_U$, respectively; $\mathbf H_r$ and $\mathbf H_U$ are the corresponding state Jacobians.
$\mathbf Q$ is the measurement covariance, which can be either the UWB covariance $\mathbf Q_d$ or the camera covariance $\mathbf Q_c$, depending the on measurement type. This system tightly couples the IMU, camera, and UWB, allowing for the joint estimation of both the robot state and the UWB state, where the robot's localization aids UWB calibration through a unified estimator.
Nevertheless, the performance of the EKF-based method is highly sensitive to two key factors: the initial UWB parameters and the accuracy of the UWB model. If UWB measurements are naively fused with incorrect initial guesses or an inaccurate model, it can degrade robot localization performance and adversely affect UWB calibration results.
To better illustrate this, we conduct numerical tests (cf. {\color{red} Sec. \ref{Sim}}) to explore how these factors affect the overall system performance, which motivates the proposed SKF-based framework for the UWB calibration.

During the UWB calibration, the robot localization provided by VINS is generally accurate as long as the robot's motion is minimal without aggressive maneuvers, whereas the UWB parameters are uncalibrated and may not be sufficiently accurate.
In such cases, using UWB measurements to update the state with poorly initialized parameters—such as an incorrect anchor position—can degrade localization accuracy and thus impact UWB calibration. This phenomenon is confirmed by the simulations described in {\color{red} Sec. \ref{Sim}}. We observe that the initialized anchor positions can exhibit significant errors in the z-direction if the robot does not undergo substantial vertical motion changes during the initialization process. This initialization error can further lead to additional localization inaccuracies when the UWB measurements are used to update the state, causing the refinement step to fail.

\noindent\textbf{Accuracy of the UWB model:} The accuracy of UWB measurements can be influenced by several factors, such as anchor configurations \cite{ZW2021} and the presence of obstacles \cite{KD2018}. Although the measurement model in \eqref{eq_uwb} 
shows satisfactory performance, it still imperfect that may not accurately capture errors across different scenarios. Deriving a universally accepted model and determining the noise level is a nontrivial task. To test the performance with different noise levels and model parameters, we conduct extensive Monte-Carlo simulations as described in {\color{red} Sec. \ref{Sim}}. From Table \ref{Tab1}, we observe that both robot localization and UWB calibration can only be improved when the noise parameters are properly tuned and the model is accurately defined. If the noise parameters are tuned too low compared to the true noise value, fusing the UWB measurements to update the state will result in inconsistent estimation, ultimately degrading localization accuracy even with a good initialized UWB parameters.

\subsection{SKF-based Update}\label{SKF}
As discussed in the previous section, tightly-coupled VINS-aided UWB calibration heavily relies on the accuracy of the initialized UWB parameters and the selection of the measurement model. However, in practice, both of these factors are highly susceptible to environmental influences, making it challenging to achieve optimal results simultaneously. Therefore, to achieve accurate UWB calibration while ensure the performance of robot localization to avoid potential inaccuracies, we propose to leverage the Schmidt-Kalman filter (SKF) \cite{EK2020}. The SKF updates only a selected subset of state variables while keeping the estimates of other variables fixed. We encourage the readers to refer to \cite{EK2020} for more detailed introduction. However, it still maintains consistency by accurately tracking all correlations.

Specifically, during the UWB update, we only update the UWB state and set the gain of the active MSCKF state to zero based on the SKF.
\begin{align}
\hat{\mathbf x}_{k+1|k+1}&=\hat{\mathbf x}_{k+1|k}\boxplus\begin{bmatrix}
\mathbf 0 \\ \mathbf K_U  
\end{bmatrix} \mathbf r\nonumber\\
{\mathbf P}_{k+1|k+1}&={\mathbf P}_{k+1|k}-
\begin{bmatrix}
\mathbf 0&\Delta \mathbf P_{rU}\\
\Delta \mathbf P_{rU}^\top&\Delta \mathbf P_{UU}\\
\end{bmatrix}
\end{align}
with
\begin{align}
\Delta \mathbf P_{rU}&=\mathbf K_r\mathbf H_x \begin{bmatrix}
\mathbf P_{rU}^\top&\mathbf P_{UU}^\top
\end{bmatrix}^\top,\quad
\Delta \mathbf P_{UU}=\mathbf K_U \mathbf S \mathbf K_U^\top
\end{align}
where $\mathbf P_{rU}$ denotes the cross-covariance of between the active MSCKF state and the UWB state as defined in \eqref{eq_state_cov}. Obviously, the active MSCKF state and its covariance do not change during the UWB calibration, whereas the UWB state is still being updated and the correlation is still being tracked. The ranging measurements only update the UWB state and its correlation, and even with a poor initial guess of the UWB state or an inaccurate model parameters, the robot's localization will remain unaffected during the calibration. Note that the visual-measurement is still updated based on the standard EKF as illustrated in {\color{red} Sec. \ref{NA}}. To ensure the consistency of the estimator, we employ First Estimate Jacobians (FEJ) when computing the Jacobian \cite{FEJ}. The proposed SKF-based approach guarantees the performance of robot localization during the UWB calibration, thereby enhancing the robustness of the calibration process.

\section{Experiments and Validations}
\subsection{Simulated Experiments}\label{Sim}
This section presents the simulation experiments to evaluate the performance and robustness of the proposed calibration method (robust initialization and SKF-based online refinement) across various scenarios, parameter settings, and noise levels. In the simulation, we utilize the visual-inertial measurements from the \textit{Euroc datasets} \cite{Euroc}, and simulate additional UWB ranging measurements from four UWB anchors, with a measurement density of $0.10m$ and a frequency of $10Hz$. The bias terms $(\beta, \gamma)$ are set to $(0.9, -0.3 \text{m})$. The used simulation platform is provided from \cite{SK2018}. The simulation platform used in this work is provided by \cite{SK2018}.

\begin{figure}
    \vspace{-1em}
    \centering
    \begin{subfigure}[h]{0.23\textwidth}
            \centering
        \includegraphics[width=\textwidth]{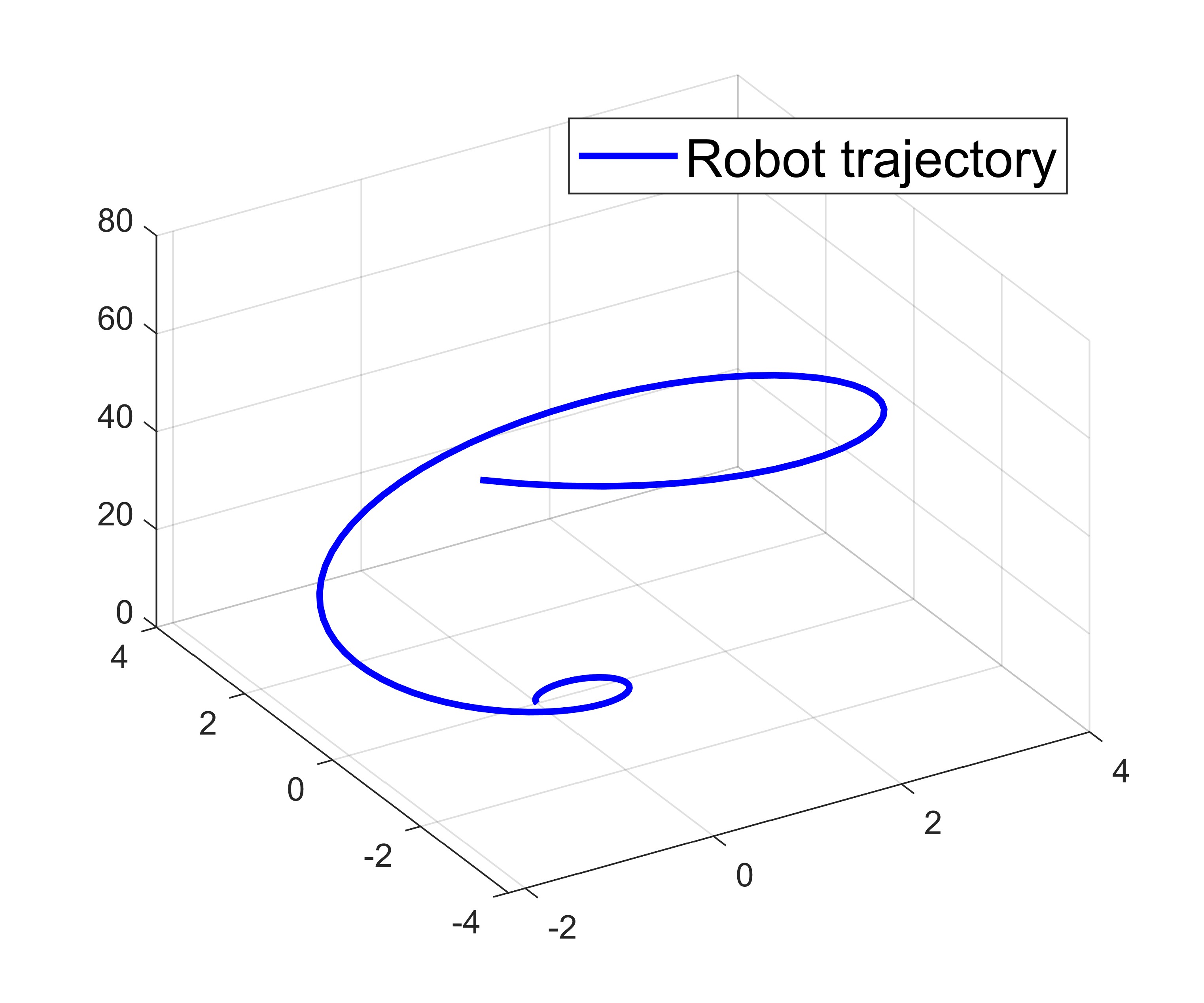}
        \caption{\small Simulated trajectory}
        \label{Fig_sim_traj1}
    \end{subfigure}
    \begin{subfigure}[h]{0.23\textwidth}
        \centering
        \includegraphics[width=\textwidth]{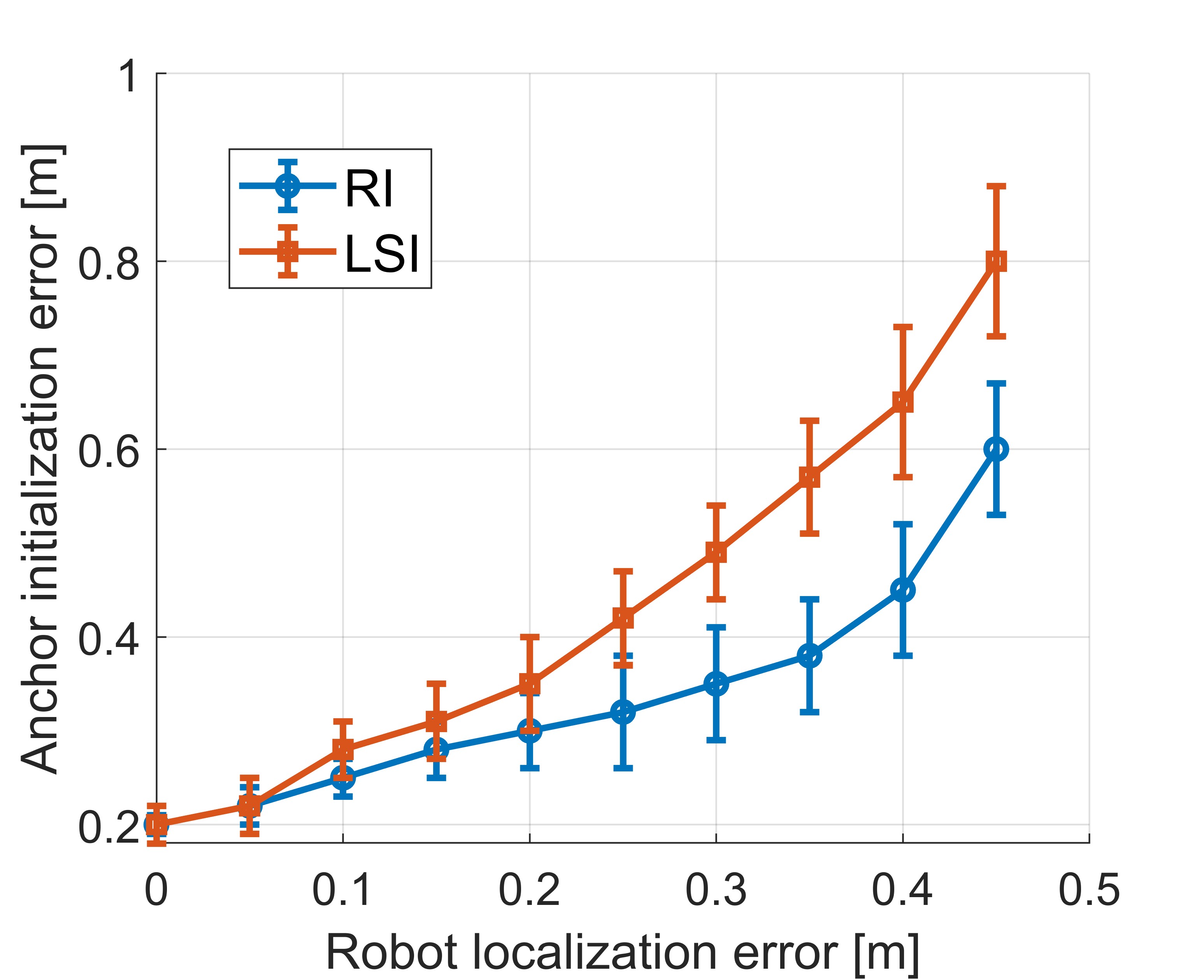}
        \caption{\small Anchor initialization errors}
        \label{Fig_sim_traj1_}
    \end{subfigure}
    \caption{Comparison of RI and LSI under various robot localization errors.}
    \vspace{-1ex}
    \label{Fig_init_cp}
\end{figure}

\vspace{1.0ex}
\noindent\textbf{Robust initialization:}
We first evaluate the initialization performance under varying levels of localization uncertainty. Monte Carlo simulations were conducted in \textit{Matlab} by generating UWB measurements and random UAV trajectories with different localization errors $\sigma_r$. The anchor position is $\begin{bmatrix} 10& 10& 10 \end{bmatrix}^\top$. From Fig.~\ref{Fig_init_cp}, when the robot doesn't have significant localization error due to it's small and smooth motion, the robust initialization (RI) method does not show significant improvement compared to standard least squares initialization (LSI) method. However, as the localization error increases, the error of the LSI method grows rapidly, whereas the proposed RI method demonstrates robustness against localization errors. Similar results are observed in the simulation experiments based on the \textit{Euroc dataset}. The proposed RI method outperforms LSI in more challenging datasets that feature aggressive robot maneuvers.

\begin{figure*}
    \centering
    \begin{subfigure}[h]{0.24\textwidth}
            \centering
        \includegraphics[width=\textwidth]{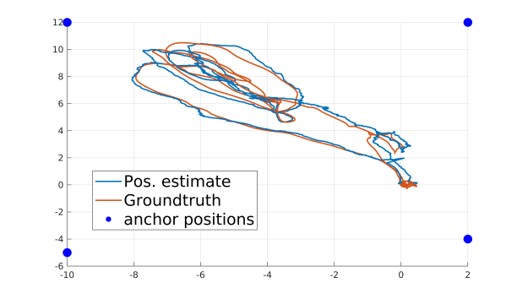}
        \caption{}
        \label{Fig_traj_re1}
    \end{subfigure}
    \begin{subfigure}[h]{0.24\textwidth}
        \centering
        \includegraphics[width=\textwidth]{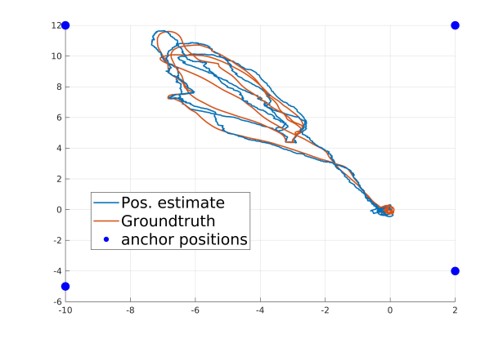}
        \caption{}
        \label{Fig_traj_re2}
    \end{subfigure}
    \begin{subfigure}[h]{0.23\textwidth}
        \centering
        \includegraphics[width=\textwidth]{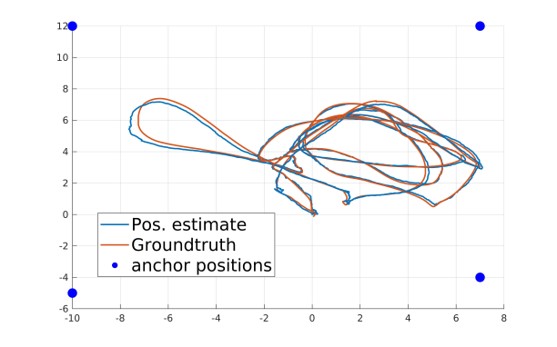}
        \caption{}
        \label{Fig_traj_re3}
    \end{subfigure}
    \begin{subfigure}[h]{0.23\textwidth}
        \centering
        \includegraphics[width=\textwidth]{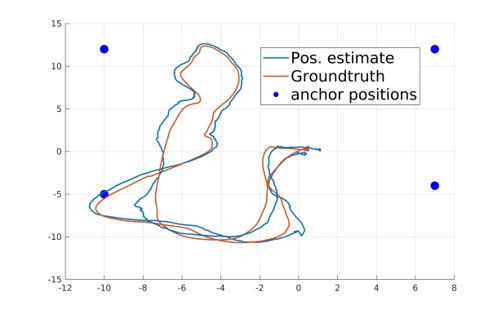}
        \caption{}
        \label{Fig_traj_re4}
    \end{subfigure}
    \caption{Estimated trajectories for all trajectory sequences of the \textit{EuRoC MH} dataset.}
    \vspace{-1ex}
    \label{Fig_est_traj}
\end{figure*}

\begin{figure}
	\centering
	\begin{subfigure}[h]{0.23\textwidth}
            \centering
		\includegraphics[width=\textwidth]{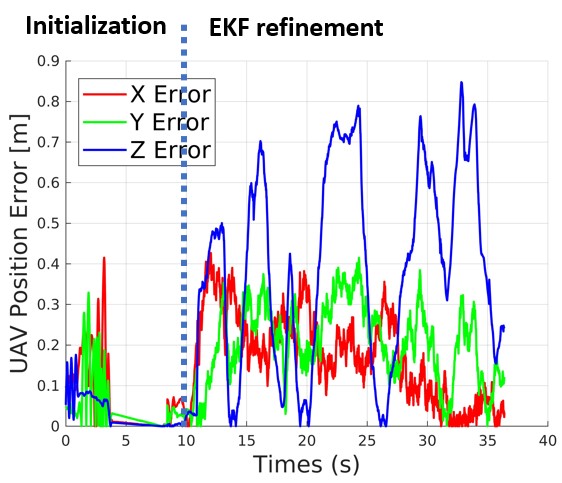}
		\caption{\small EKF}
		\label{Fig_sim_traj1}
	\end{subfigure}
	\begin{subfigure}[h]{0.23\textwidth}
		\centering
		\includegraphics[width=\textwidth]{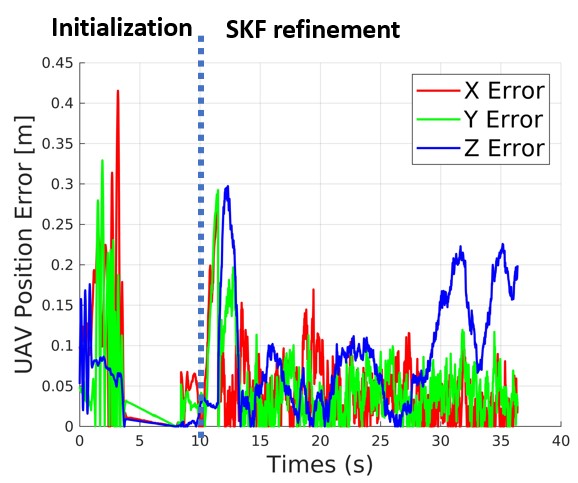}
		\caption{\small SKF}
		\label{Fig_sim_traj1_}
	\end{subfigure}
	\caption{Comparison of SKF vs. EKF refinement}
 \vspace{-1ex}
    \label{Fig_sim_traj}
\end{figure}

\vspace{1.0ex}
\noindent\textbf{SKF-based online refinement:} 
Based on the setting from Fig.~\ref{Fig_exp_diag}, we first present the results from \textit{MH01, Euroc datasets} as a representative example to analyze the algorithm's performance. As shown in Fig.~\ref{Fig_sim_traj}, due to the robot's limited motion in the z-axis during the UWB initialization, the UWB initial estimate has a large error in that direction (as we analyzed in (cf. {\color{red} Sec. \ref{FIM}})). Thus, the EKF exhibits significant localization errors in the second stage, as it directly fuses the UWB ranging measurements to update the UAV state using an imperfect initial UWB estimate from the initialization step, which negatively impacts both localization accuracy and calibration. In contrast, the proposed SKF only uses UWB measurements to update the UWB state during calibration, ensuring that localization is not affected by inaccurate initial UWB values and improving the calibration as well. We also present the estimation results for all trajectories, as illustrated in Figure \ref{Fig_est_traj}.

\begin{table}[htbp]
\centering
\caption{Comparison of SKF and EKF under various initial error $\bm\sigma_I$ and noise parameter $\bm\sigma_U$ ({${^G \widetilde{\mathbf p}}_a$} denotes the anchor positioning error) }
\begin{tabular}{ccccc}
\toprule
$\bm\sigma_I (m)$ & $\bm \sigma_U$ & & {PRMSE (m)} & {${^G \widetilde{\mathbf p}}_a$ (m)}\\
\hline
\multirow{2}{*}{0.1}&\multirow{2}{*}{1.0} & SKF & 0.452 & 0.312 \\
                  &                   & EKF & \textbf{0.379} & \textbf{0.271} \\
\arrayrulecolor[gray]{0.8}
\hline
\multirow{2}{*}{0.3}&\multirow{2}{*}{1.0} & SKF & 0.487 & \textbf{0.346} \\
                  &                   & EKF & \textbf{0.466} & 0.349 \\
\arrayrulecolor[gray]{0.8}
\hline
\multirow{2}{*}{0.3}&\multirow{2}{*}{0.9} & SKF & \textbf{0.489} & \textbf{0.352} \\
                  &                   & EKF & 0.546 & 0.408 \\
\arrayrulecolor[gray]{0.8}
\hline
\multirow{2}{*}{0.5}&\multirow{2}{*}{0.9} & SKF & \textbf{0.495} & \textbf{0.422} \\
                  &                   & EKF & 0.679 & 0.771 \\
\arrayrulecolor{black}
\bottomrule
\vspace{-1ex}
\label{Tab1}
\end{tabular}
\end{table}

We further evaluate the performance of the SKF-based method across different UWB initial values and model accuracies. We introduce additional error $\bm\sigma_I$ to the anchor parameters to simulate varying initialization errors and adjust the UWB noise by multiplying it with a factor, $\bm\sigma_U$ (with $\bm\sigma_U = 1$ indicating accurately tuned noise). As shown in Table \ref{Tab1}, we observe that the EKF outperforms the proposed SKF when the model is accurate and the UWB parameters have a precise initial guess, as it incorporates the ranging measurements to jointly update both the robot state and the UWB state.
Nevertheless, achieving both conditions simultaneously is difficult in real-world scenarios. Extensive experimental results still demonstrate that a imperfect initial guess of the UWB parameter can degrade the localization accuracy of the EKF-based method, which consequently hurts the calibration results, while the SKF-based method maintains a robust localization performance across different initial values. Since the SKF only uses the ranging measurements to update the UWB state without correcting the active MSCKF state, the accuracy of the initial guess has minimal effect on localization performance. This effectively reduces the dependency on the precision of the initial guess.

\subsection{Real-world Experiments}
We further validate the proposed method by a real-world dataset, \textit{NTU-VIRAL} \cite{NTUviral}, which provides measurement data of IMU, stereo camera, and UWB ranging measurements from three anchors. 

\noindent\textbf{Experimental settings:} As illustrated in Fig. ~\ref{Fig_exp_diag}, the trajectory data of each dataset is divided into three parts for testing (as we don't need the complete robot trajectory that is too for calibration). The first two parts are used to evaluate the two stages of the proposed calibration framework: robust initialization and SKF-based online refinement. The third stage, UWB-aided UAV localization, is performed based on the completion of the calibration process to evaluate its effectiveness. In other word, in each running of the trajectory sequence, the robot will first perform the calibration process then incorporate the calibrated UWB 
to update the estimate. Since the the focus of this paper is the robot-aided calibration process, here we mainly present the calibration results (\textit{specifically} the robot localization and anchor estimation during the calibration process), and compared the full algorithm results (\textit{RI+SKF}) with \textit{FEJ-VIRO} \cite{Jia2022} (referred to as VIRO for simplicity), and \textit{RI(proposed)+EKF}. The robot localization performance during calibration is evaluated using the root mean square error (RMSE).

\noindent\textbf{Calibration results analysis:} Table \ref{Tab2} presents the trajectory accuracy and anchor estimation accuracy for all methods. Since we only used a portion of the trajectory for calibration, the results are based solely on that segment of the trajectory. We observe that the proposed RI method improves anchor estimation across all trajectory trials.
Additionally, \textit{RI+EKF} achieves the best performance in both localization and calibration when the robot avoids aggressive motion and its movement is not restricted to a single plane during the initialization phase, with VIRO showing almost comparable results. This is because the robot’s smooth motion ensures an accurate UWB initialization process. With a reliable initial UWB estimate, both methods update the robot and UWB states jointly using EKF during the refinement phase, leading to more accurate state estimation and calibration than SKF, which only updates the UWB state during calibration. However, this approach significantly limits the robot’s movement. If the robot experiences larger drifts during initialization, it compromises the accuracy of the process. The joint EKF update of the robot and UWB states, when based on an inaccurate initial UWB estimate, further degrades both localization and calibration in the EKF-based \textit{VIRO} and \textit{RI+EKF} methods. In such cases, the SKF method proves more robust, as it better calibrates the UWB, ensuring that localization is less affected by a poor initial UWB estimate. 

\begin{table}[htbp]
\centering
\caption{Performance of the proposed calibration method in \textit{NTU-VIRAL dataset} (on calibration trajectory segments only)}
\begin{tabular}{ccccccc}
\toprule
Dataset &Algorithms &  {PRMSE (m)} & {ORMSE (deg)} & {${^G \widetilde{\mathbf p}}_a$ (m)}\\
\hline
\multirow{4}{*}{eee01}
&VIRO & 0.351 & 5.725 & 0.485\\
&RI+EKF & 0.300& 5.723& 0.408\\
&RI+SKF & \textbf{0.287}& \textbf{5.626}& \textbf{0.346}\\
\arrayrulecolor[gray]{0.8}
\hline
\multirow{4}{*}{eee03}
&VIRO & 0.329& 4.867& 0.425\\
&RI+EKF & \textbf{0.289} & \textbf{4.865}& \textbf{0.422}\\
&RI+SKF & 0.325& 4.871& 0.429\\
\arrayrulecolor[gray]{0.8}
\hline
\multirow{4}{*}{nya01}
&VIRO & 0.326& 4.652& 0.412\\
&RI+EKF & 0.319& 4.674& 0.388\\
&RI+SKF & \textbf{0.301}& \textbf{4.600}& \textbf{0.375}\\
\arrayrulecolor{black}
\bottomrule
\label{Tab2}
\end{tabular}
\end{table}

\section{Conclusion}
This paper presents a robust online calibration framework for UWB-aided VINS systems, addressing the limitations of previous methods that were sensitive to localization errors and initial anchor guesses. By incorporating robot localization uncertainty into the UWB initialization process and employing a tightly-coupled SKF-based online refinement, the proposed method significantly enhances calibration accuracy and robustness. Extensive simulated and real-world experiments validate the effectiveness of the approach. In the future, we plan to extend the framework proposed in this paper to develop active planning-based calibration algorithms. The focus will be on studying the impact of robot trajectories, localization errors, and anchor placement on the calibration results.

\bibliographystyle{IEEEtran}

\bibliography{Ref_main} 

\begin{thebibliography}{10}
\providecommand{\url}[1]{#1}
\csname url@samestyle\endcsname
\providecommand{\newblock}{\relax}
\providecommand{\bibinfo}[2]{#2}
\providecommand{\BIBentrySTDinterwordspacing}{\spaceskip=0pt\relax}
\providecommand{\BIBentryALTinterwordstretchfactor}{4}
\providecommand{\BIBentryALTinterwordspacing}{\spaceskip=\fontdimen2\font plus
\BIBentryALTinterwordstretchfactor\fontdimen3\font minus \fontdimen4\font\relax}
\providecommand{\BIBforeignlanguage}[2]{{%
\expandafter\ifx\csname l@#1\endcsname\relax
\typeout{** WARNING: IEEEtran.bst: No hyphenation pattern has been}%
\typeout{** loaded for the language `#1'. Using the pattern for}%
\typeout{** the default language instead.}%
\else
\language=\csname l@#1\endcsname
\fi
#2}}
\providecommand{\BIBdecl}{\relax}
\BIBdecl

\bibitem{OP2020}
P.~Geneva, K.~Eckenhoff, W.~Lee, Y.~Yang, and G.~Huang, ``Openvins: A research platform for visual-inertial estimation,'' in \emph{2020 IEEE International Conference on Robotics and Automation (ICRA)}, 2020, pp. 4666--4672.

\bibitem{VM}
T.~Qin, P.~Li, and S.~Shen, ``Vins-mono: A robust and versatile monocular visual-inertial state estimator,'' \emph{IEEE Transactions on Robotics}, vol.~34, no.~4, pp. 1004--1020, 2018.

\bibitem{Jia2022}
S.~Jia, Y.~Jiao, Z.~Zhang, R.~Xiong, and Y.~Wang, ``Fej-viro: A consistent first-estimate jacobian visual-inertial-ranging odometry,'' in \emph{2022 IEEE/RSJ International Conference on Intelligent Robots and Systems (IROS)}, 2022, pp. 1336--1343.

\bibitem{Hc2024}
C.~Hu, P.~Huang, and W.~Wang, ``Tightly coupled visual-inertial-uwb indoor localization system with multiple position-unknown anchors,'' \emph{IEEE Robotics and Automation Letters}, vol.~9, no.~1, pp. 351--358, 2024.

\bibitem{NTM2021}
T.-M. Nguyen, M.~Cao, S.~Yuan, Y.~Lyu, T.~H. Nguyen, and L.~Xie, ``Liro: Tightly coupled lidar-inertia-ranging odometry,'' in \emph{2021 IEEE International Conference on Robotics and Automation (ICRA)}, 2021, pp. 14\,484--14\,490.

\bibitem{Hj2023}
J.~Hu, Y.~Li, Y.~Lei, Z.~Xu, M.~Lv, and J.~Han, ``Robust and adaptive calibration of uwb-aided vision navigation system for uavs,'' \emph{IEEE Robotics and Automation Letters}, vol.~8, no.~12, pp. 8247--8254, 2023.

\bibitem{BJ2021}
J.~Blueml, A.~Fornasier, and S.~Weiss, ``Bias compensated uwb anchor initialization using information-theoretic supported triangulation points,'' in \emph{2021 IEEE International Conference on Robotics and Automation (ICRA)}, 2021, pp. 5490--5496.

\bibitem{NT2020}
T.~H. Nguyen, T.-M. Nguyen, and L.~Xie, ``Tightly-coupled single-anchor ultra-wideband-aided monocular visual odometry system,'' in \emph{2020 IEEE International Conference on Robotics and Automation (ICRA)}, 2020, pp. 665--671.

\bibitem{NT2022}
T.~H. Nguyen and L.~Xie, ``Estimating odometry scale and uwb anchor location based on semidefinite programming optimization,'' \emph{IEEE Robotics and Automation Letters}, vol.~7, no.~3, pp. 7359--7366, 2022.

\bibitem{WC2017}
C.~Wang, H.~Zhang, T.-M. Nguyen, and L.~Xie, ``Ultra-wideband aided fast localization and mapping system,'' in \emph{2017 IEEE/RSJ International Conference on Intelligent Robots and Systems (IROS)}, 2017, pp. 1602--1609.

\bibitem{PF2017}
F.~J. Perez-Grau, F.~Caballero, L.~Merino, and A.~Viguria, ``Multi-modal mapping and localization of unmanned aerial robots based on ultra-wideband and rgb-d sensing,'' in \emph{2017 IEEE/RSJ International Conference on Intelligent Robots and Systems (IROS)}, 2017, pp. 3495--3502.

\bibitem{ZJ2022}
J.-R. Zhan and H.-Y. Lin, ``Improving visual inertial odometry with uwb positioning for uav indoor navigation,'' in \emph{2022 26th International Conference on Pattern Recognition (ICPR)}, 2022, pp. 4189--4195.

\bibitem{YB2021}
B.~Yang, J.~Li, and H.~Zhang, ``Uvip: Robust uwb aided visual-inertial positioning system for complex indoor environments,'' in \emph{2021 IEEE International Conference on Robotics and Automation (ICRA)}, 2021, pp. 5454--5460.

\bibitem{SSL2021}
S.~Shin, E.~Lee, J.~Choi, and H.~Myung, ``Mir-vio:mutual information residual-based visual inertial odometry with uwb fusion for robust localization,'' in \emph{2021 21st International Conference on Control, Automation and Systems (ICCAS)}, 2021, pp. 91--96.

\bibitem{ZS20222}
S.~Zheng, Z.~Li, Y.~Liu, H.~Zhang, P.~Zheng, X.~Liang, Y.~Li, X.~Bu, and X.~Zou, ``Uwb-vio fusion for accurate and robust relative localization of round robotic teams,'' \emph{IEEE Robotics and Automation Letters}, vol.~7, no.~4, pp. 11\,950--11\,957, 2022.

\bibitem{ZW2021}
W.~Zhao, J.~Panerati, and A.~P. Schoellig, ``Learning-based bias correction for time difference of arrival ultra-wideband localization of resource-constrained mobile robots,'' \emph{IEEE Robotics and Automation Letters}, vol.~6, no.~2, pp. 3639--3646, 2021.

\bibitem{JS2023}
S.~Jia, R.~Xiong, and Y.~Wang, ``Distributed initialization for visual-inertial-ranging odometry with position-unknown uwb network,'' in \emph{2023 IEEE International Conference on Robotics and Automation (ICRA)}, 2023, pp. 6246--6252.

\bibitem{DGS2023}
G.~Delama, F.~Shamsfakhr, S.~Weiss, D.~Fontanelli, and A.~Fomasier, ``Uvio: An uwb-aided visual-inertial odometry framework with bias-compensated anchors initialization,'' in \emph{2023 IEEE/RSJ International Conference on Intelligent Robots and Systems (IROS)}, 2023, pp. 7111--7118.

\bibitem{MSCKF}
A.~I. Mourikis and S.~I. Roumeliotis, ``A multi-state constraint kalman filter for vision-aided inertial navigation,'' in \emph{Proceedings 2007 IEEE International Conference on Robotics and Automation}, 2007, pp. 3565--3572.

\bibitem{SK2018}
K.~Sun, K.~Mohta, B.~Pfrommer, M.~Watterson, S.~Liu, Y.~Mulgaonkar, C.~J. Taylor, and V.~Kumar, ``Robust stereo visual inertial odometry for fast autonomous flight,'' \emph{IEEE Robotics and Automation Letters}, vol.~3, no.~2, pp. 965--972, 2018.

\bibitem{boyd2004}
S.~Boyd and L.~Vandenberghe, \emph{Convex Optimization}.\hskip 1em plus 0.5em minus 0.4em\relax Cambridge University Press, 2004.

\bibitem{zzSPLY2025}
\BIBentryALTinterwordspacing
Y.~Zhou, ``Supplementary materials: A consistent and tightly-coupled visual-inertial-ranging odometry on lie groups,'' 2025. [Online]. Available: \url{https://mason.gmu.edu/~xwang64/papers/rcvin_supp.pdf}
\BIBentrySTDinterwordspacing

\bibitem{WZ2008}
Z.~Wang and G.~Dissanayake, ``Observability analysis of slam using fisher information matrix,'' in \emph{2008 10th International Conference on Control, Automation, Robotics and Vision}, 2008, pp. 1242--1247.

\bibitem{Kay1993}
S.~M. Kay, \emph{Fundamentals of Statistical Signal Processing: Estimation Theory}.\hskip 1em plus 0.5em minus 0.4em\relax Upper Saddle River, NJ, USA: Prentice-Hall, Inc., 1993.

\bibitem{KD2018}
D.-H. Kim, G.-R. Kwon, J.-Y. Pyun, and J.-W. Kim, ``Nlos identification in uwb channel for indoor positioning,'' in \emph{2018 15th IEEE Annual Consumer Communications \& Networking Conference (CCNC)}, 2018, pp. 1--4.

\bibitem{EK2020}
K.~Eckenhoff, P.~Geneva, N.~Merrill, and G.~Huang, ``Schmidt-ekf-based visual-inertial moving object tracking,'' in \emph{2020 IEEE International Conference on Robotics and Automation (ICRA)}, 2020, pp. 651--657.

\bibitem{FEJ}
G.~P. Huang, A.~I. Mourikis, and S.~I. Roumeliotis, ``A first-estimates jacobian ekf for improving slam consistency,'' in \emph{Experimental Robotics}, O.~Khatib, V.~Kumar, and G.~J. Pappas, Eds.\hskip 1em plus 0.5em minus 0.4em\relax Berlin, Heidelberg: Springer Berlin Heidelberg, 2009, pp. 373--382.

\bibitem{Euroc}
M.~Burri, J.~Nikolic, P.~Gohl, T.~Schneider, J.~Rehder, S.~Omari, M.~W. Achtelik, and R.~Siegwart, ``The euroc micro aerial vehicle datasets,'' \emph{The International Journal of Robotics Research}, vol.~35, no.~10, pp. 1157--1163, 2016.

\bibitem{NTUviral}
T.-M. Nguyen, S.~Yuan, M.~Cao, Y.~Lyu, T.~H. Nguyen, and L.~Xie, ``Ntu viral: A visual-inertial-ranging-lidar dataset, from an aerial vehicle viewpoint,'' \emph{The International Journal of Robotics Research}, vol.~41, no.~3, pp. 270--280, 2022.

\end{thebibliography}

\end{document}